%% file: tacl2018v2-template.tex
\documentclass[11pt,a4paper]{article}
\usepackage{times,latexsym}
\usepackage{url}
\usepackage[T1]{fontenc}
\usepackage{subcaption}
\usepackage{graphicx}
    \usepackage[acceptedWithA]{tacl2018v2}
\usepackage[]{tacl2018v2}

\usepackage{xspace,mfirstuc,tabulary}

\title{Sensing Ambiguity in Henry James' "The Turn of the Screw"}

\author{Victor Makarenkov\\
	    {\tt vitiokm@gmail.com}
	  \And
	Yael Segalovitz\\
  	Ben-Gurion University of the Negev\\
  	Beer Sheva, Israel\\
  {\tt yaelsega@bgu.ac.il}}

\date{}

\begin{document}
\maketitle
\begin{abstract}
Fields such as the philosophy of language, continental philosophy, and literary studies have long established that human language is, at its essence, ambiguous and that this quality, although challenging to communication, enriches language and points to the complexity of human thought. On the other hand, in the NLP field there have been ongoing efforts aimed at disambiguation for various downstream tasks. This work brings together computational text analysis and literary analysis to demonstrate the extent to which ambiguity in certain texts plays a key role in shaping meaning and thus requires analysis rather than elimination. We revisit the discussion, well known in the humanities, about the role ambiguity plays in Henry James’ 19th century novella, “The Turn of the Screw.” We model each of the novella’s two competing interpretations as a topic and computationally demonstrate that the duality between them exists consistently throughout the work and shapes, rather than obscures, its meaning. We also demonstrate that cosine similarity and word mover’s distance are sensitive enough to detect ambiguity in its most subtle literary form, despite doubts to the contrary raised by literary scholars. Our analysis is built on topic word lists and word embeddings from various sources. We first claim, and then empirically show, the interdependence between computational analysis and close reading performed by a human expert.

\end{abstract}
\input{macros}

\section{Introduction}

In the NLP field, there is an assumption that natural language is both the easiest and the main way for humans to communicate, yet it is ambiguous and accompanied by the challenge of correct interpretation. Consequently, disambiguation has always been the focus of NLP research. \citet{aina-etal-2019-putting} examined LSTM networks and language models that can deal with lexical ambiguity. \citet{elkahky-etal-2018-challenge} created a dataset with non-trivial noun-verb ambiguity and challenged the development of new part-of-speech (POS) taggers. Early \cite{chen-etal-1999-resolving} and more recent \cite{zhao-etal-2018-addressing} research has dealt with word disambiguation in the process of statistical machine translation. Other disambiguation research used knowledge base and probabilistic graph based methods \cite{hoffart-etal-2011-robust, moro-etal-2014-entity, cucerzan-2007-large}. More recent work employed deep neural networks \cite{ganea-hofmann-2017-deep, raganato-etal-2017-word} for the disambiguation task.

Historically, literary studies and related fields in the humanities have held a different view of ambiguity, seeing it as an important facet of language, which is, in fact, enhanced in aesthetically artful works, that is, literary works often make use of the ambiguity of language in order to communicate complex ideas and manipulate textual forms, as well as to challenge the reader or to draw the reader in. In this work, we focus on Henry James' novella, "The Turn of the Screw" \cite{Screw1898}. Literary scholars have demonstrated that novella is structured around systematic ambiguity that is meant to instigate hesitation in the reader (see Section \ref{section:discusion}). In this paper, we show that NLP can detect and numerically corroborate the existence of such ambiguity and can demonstrate the extent to which this ambiguity is systematic and rhythmic. We do not attempt to disambiguate the novella. Instead, we computationally analyze and understand its ambiguous meaning.

"The Turn of the Screw" is the author's most popular work. In fact, as Peter G. Beidler points out, it is “one of the most widely discussed pieces of fiction ever written” \cite{james1995turn}. For over a century this novella has generated heated scholarly debate and spurred numerous articles, books \cite{norton, beidler1995critical}, and student papers, as well as films (e.g., \textit{The Others}, 2001 and \textit{The Turning}, 2020) and book adaptations \cite{oates1994haunted} that present authors' interpretations of the tale.

As discussed below (see Section \ref{section:duality}), at the center of James' novella there is a fundamental ambiguity which prominent literary scholars view as an aspect of the work's strength \cite{felman1977turning}. In the words of \citet{bradleithouser} “All such attempts to 'solve' the book, however admiringly tendered, unwittingly work toward its diminution... Its profoundest pleasure lies in the beautifully fussed over way in which James refuses to come down on either side... the book becomes a modest monument to the bold pursuit of ambiguity".

However, literary scholars have voiced doubts about the ability of computational analyses to account for such nuanced literary ambiguity. That is, since the early aughts, literary studies have shown an increasing interest in computational analysis, resulting in the establishment of the emerging field of the digital humanities \cite{kirschenbaum2007remaking, gold2012debates, janicke2015close}. Yet, literary scholars are still generally skeptical about computational literary analysis, especially with regard to its ability to replicate a trained reader’s sensitivity to linguistic and semantic nuance \cite{hammond-etal-2013-tale, kopec2016digital, eyers2013perils}. Other critics have commented on the unlikelihood of a scholar attaining sufficient expertise in both computational and literary analysis to be able to perform adequately. Current research demonstrates the benefits and advantages of interdisciplinary collaboration between NLP and literary scholars, leading to the reasonable conclusion that close reading \cite{segalovitz2019william} and computational analysis are not mutually exclusive but rather are \textit{interdependent} \cite{kopec2016digital}. With the guidance of a close reader, computational analysis can detect, examine and calculate the most subtle ambiguity, and with the aid of computational analysis, literary scholars can gain visual and quantitative insight into complex literary phenomena.

In this work, we computationally analyze "The Turn of the Screw" and its shift between two possible main interpretations.  We build our analysis on standard NLP techniques such as Latent Dirichlet Allocation (LDA) \cite{LDA}, Kullback-Leibrer (KL) divergence \cite{kullback1951information}, word embeddings and punctuation ratio techniques. We demonstrate the building of a list of words for each of the two possible main interpretations of "The Turn of the Screw" and use two different metrics to quantify the dual interpretation of the text. Our results show that man-machine collaboration results in the best consonant demonstration and explanation of textual ambiguity. In addition, the use of word embeddings reveals the productivity of a synthesis between close reading and distant reading \cite{moretti2013distant}. Put otherwise, we show how a trained reader’s work with the details of a single text can be enriched by the computational analysis of a multitude of texts, in this case, James’ entire oeuvre.

This paper is organized as follows: Section \ref{section:related} surveys the related and background work. Section \ref{section:plot} summarizes the novella's plot. Section \ref{section:duality} presents the two main possible interpretations of the plot. The computational analysis of the ambiguity is presented in Section \ref{section:analysis}. Section \ref{section:content} examines the timeline of the novella's plot. The discussion about the current work is placed in Section \ref{section:discusion}. Section \ref{section:conclusions} is dedicated to final conclusions and future work. 
       
\section{Related Work}
\label{section:related}
Disambiguation and content analysis is a broad domain within NLP research. Content analysis tasks relevant to this area of research are those in which the subject of interest might not directly stated in the text. There are multiple examples of tasks in which a detection of indirectly specified matter is made. One example is the sentiment analysis \cite{liu2012sentiment, pak2010twitter} which is a well established research field where a typical task is to classify the text as a negative, positive, or neutral sentiment. \citet{yu2017refining} and \citet{giatsoglou2017sentiment} utilized word embeddings for the sentiment analysis task. Another example is the analysis of community question answering (CQA) archives. \citet{harper09facts} first studied the task of identifying informational and conversational questions within CQA archives; in the questions, the users do not specify their explicit intent - should a conversation be started, or just a plain need for precise information. Recently, this task was revisited by \citet{10.1145/3159652.3159733} whose improved performance on this task was achieved due to the use of word embeddings. Another interesting example is connected to news outlets political bias detection. \citet{makarenkov2019implicit} presented an analysis of political perspectives and leanings that implicitly arise in contemporary online media sources. They used both pre-trained word embeddings and an LSTM classifier to demonstrate the presence of political perspectives in presumably neutral European and American news sources. 
A more artistic example is the case of movies' overviews analysis. \citet{gorinski-lapata-2018-whats} generated movie overviews with an LSTM decoder. They exploited movie scripts and other natural language texts about the plots, genres and artistic styles of the movies to compose the movie overview's narrative.
A central disambiguation case is present as a part of machine translation task. \citet{mascarell-etal-2015-detecting} proposed detecting document-level context features in order to support disambiguation in the task of machine translation. 
Another example we mention is of \citet{mao-etal-2018-word}, who used continuous bag of words (CBOW) and skip gram negative sampling (SGNS) \cite{mikolov2013efficient} embedding architectures combined with WordNet \cite{miller1998wordnet} for metaphor interpretation.

The use of pre-trained off-the-shelf word embeddings received an even greater boost after \citet{devlin2018bert} introduced the pre-trained BERT transformer \cite{vaswani2017attention}. This dense representation achieved an enhanced performance in various NLP tasks \cite{goldberg2019assessing} and was popularized with its pre-trained models in the Tensorflow \cite{tensorflow2015-whitepaper} Hub platform. 

 \citet{hammond-etal-2013-tale} explained the difference between literary and computational attitude toward implicit ambiguity as follows:  while NLP researchers often see an ambiguity or polysemous text as problem that has to be solved, in literary analysis the ambiguity must remain and be present as such, as intended by the writer, without any goal to solve it. We hypothesize that this might be the reason why computational literary analysis works are very scarce \cite{roque-2012-towards}.

In our work, we did not exploit word embeddings to solve the ambiguity. Rather we exploited them to computationally shed light on the existence and presence of ambiguity and two different interpretations the reader might perceive when reading the novella.

\section{"The Turn of the Screw": The Plot}
\label{section:plot}

"The Turn of the Screw" opens with a story about the main story. The prologue, told from the first-person perspective, shares the journey the story has traveled to reach its narrator, who sits “round the fire” among a group of friends at a remote old house on Christmas in 1890s England (p. 3). The reader learns that the story was first told in complete confidence to Douglas, a guest at the house, by a governess he met about 40 years earlier. The governess bequeathed her diary which contained written documentation of the event to Douglas; upon her death, he read the entry discussing the event aloud to his friends, among them the narrator who shares the story of Douglas’ recitation with the readers in the prologue.
It becomes evident, then, that questions of perspective, reliable narration, and perception will be central to the novella no less than the plot itself: readers are urged to wonder whether they should believe the story despite its thrice-removed form, doubt who in fact is telling the story, and speculate whether her/his point of view affects the events depicted. This epistemological uncertainty permeates the entire book, which is narrated in the 24 chapters following the prologue by the unnamed governess. The youngest daughter of a country parson, the governess leaves her sheltered life for the big city in order to find a job. However, after suffering a long illness that leaves her looking pale and weak, she encounters difficulties in her pursuits and finds herself accepting a somewhat dubious position offered to her by an elegant young gentleman of great means whom she immediately falls for. He invites her to come to Bly, an isolated country house, and take care of his young niece and nephew, who were placed under his care after their parents’ death in India. However, he demands that she never contact him about the children, no matter the circumstances. The governess depicts her young charges, Flora and Miles, as the most beautiful, angelic creatures. Yet, after finding out that Miles was dismissed from his boarding school for shadowy reasons unspecified in the headmaster’s letter, the governess begins to see ghosts about the house and comes to believe that the children are in danger from them. Upon hearing the governess describe the ghosts, the housekeeper, Mrs. Grose, recognizes them as Peter Quint, the uncle’s former valet, and Mrs. Jessel, the children's former governess. The pair, it seems, have had an intimate relationship, an unacceptable liaison in Victorian times, and both died in conspicuous circumstances. They have returned, the governess decides, in order to lure the children into hell, and she makes it her mission to purge the place and free the children. Though Miles and Flora fervently deny any encounter with ghostly spirits, the governess insists that they are lying, cast under the ghosts’ malevolent spell. She therefore instructs Mrs. Grose to take Flora away from Bly in order to extract a confession from Miles about his relationship with the ghost of Quint. It is here that the novella comes to its tragic end; while Miles swears to have never met the ghost, the terrified governess detects Quint at the window and physically forces Miles to confront the image. This battle ends with the death of Miles, “his little heart, dispossessed, had stopped” (p. 125). Was his death the result of a fatal shock evoked by the confessed encounter with the ghost? Was it a result of the unbearable fear accompanying the governess’ suggestion? Perhaps it was instead caused by strangulation resulting from the tight, ostensibly protective embrace of the governess(“I caught him, yes, I held him—it may be imagined with what a passion” [p. 125])? James leaves it to the reader to decide.

\section{Ambiguity in "The Turn of the Screw"}
\label{section:duality}
From around 1920 onward, persistent uncertainties  drive discussion on the novella: 1) \textbf{Are the ghosts of Bly real}, or 2) \textbf{are they a figment of the governess’ imagination?} These uncertainties give rise to other questions: Is the governess the epitome of self-sacrifice, willing to risk herself for the sake of the children, or is she an insane, possessive caretaker, forcing her charges into the realm of her hallucinations? Are the children evil wolves in sheep’s clothing or are they innocent victims? Evidence on both sides is ample; here are but a few examples for the sake of demonstration. 

\begin{enumerate}
    \item \textbf{The real existence of the ghosts} is supported by the ability of the governess to describe Quint's ghost in such detail that Mrs. Grose immediately recognizes him as the former valet, without the governess ever hearing or knowing about Quint before. In addition, in the epilogue, the narrator is told by Douglas that the governess (whom he finds charming) has become a respected and much loved governess of other small children, which seems unlikely were she to be a neurotic and unreliable person suffering from hallucinations. Finally, by the end of the novella, even Mrs. Grose admits to believing that the children are under the influence of the ghosts of the former valet and governess. We model this interpretation as the \textit{ghost topic}.
    \item \textbf{The mental instability of the governess} is corroborated by the fact that she is only one who sees the ghosts for certain according to the text; Flora and Miles deny allegations of their presence and Mrs. Grose, even when faced with the alleged apparitions, claims to see nothing. The governess is also young, has lived a sheltered life, is recuperating from a long illness, is influenced by her infatuation with the uncle, is sexually inexperienced and hence both shocked and fascinated by sexual innuendo, tends towards exaggerations and binary thinking, and has been sleep-deprived ever since her arrival at Bly as a result of her excitement and anxiety. We model this interpretation as the \textit{insanity topic}.
\end{enumerate}

While these two mutually exclusive interpretations were thoroughly examined one against the other around the mid-20th century, from the 1980s onward critics have generally shifted from an either-or understanding of the text to a broader, more inclusive understanding in which there is a room for multiple interpretations of the text \cite{brooke1976squirm, felman1977turning, rimmon1977concept}. A more recent paradigm which has since taken root, suggests that no one true answer to the question regarding the existence of the ghosts is to be found in "The Turn of the Screw". Instead, the novella is intentionally structured around ambiguity, namely, it is fashioned such that at any given moment the story lends itself to multiple interpretations, thus stimulating in the reader precisely the kind of curiosity and engagement that could keep interest in a story alive for over a century. 

\section{A Computational Analysis of the Ambiguity}
\label{section:analysis}

\begin{table*}[]
\centering
\begin{tabular}{lll}
\hline
\multicolumn{1}{c}{\textbf{Strategy}} & \multicolumn{1}{c}{\textbf{Insanity topic}} & \multicolumn{1}{c}{\textbf{Ghost topic}} \\ \hline
\textbf{\begin{tabular}[c]{@{}l@{}}Manually from \\ Wikipedia\end{tabular}} & \begin{tabular}[c]{@{}l@{}}hallucination, madness, sickness, \\ illness, dream, confusion, psychosis, \\ illusion\end{tabular} & \begin{tabular}[c]{@{}l@{}}ghost, apparition, haunt, phantom, \\ poltergeist, shade, specter, spirit, \\ spook, wraith, soul\end{tabular} \\ \hline
\textbf{\begin{tabular}[c]{@{}l@{}}Manually from \\ the novella\end{tabular}} & \begin{tabular}[c]{@{}l@{}}fancies, fancy, fancied, anxious, \\ nervous, nerves, shock, shaken,\\  spell, sane, sanity, insane, exciting, \\ distress, impression\end{tabular} & \begin{tabular}[c]{@{}l@{}}visitation, visitant, visitor, strange, \\ stranger, queer, apparition, \\ monstrous, evil, unnatural\end{tabular} \\ \hline
\textbf{\begin{tabular}[c]{@{}l@{}}Computationally from \\ Wikipedia\end{tabular}} & \begin{tabular}[c]{@{}l@{}}mental, illnesses, ill, sick, \\ diagnosed, suffering, insanity, \\ ailment, disorder, debilitating\end{tabular} & \begin{tabular}[c]{@{}l@{}}ghost, demon, beast, alien, creature, \\ supernatural, haunted, mysterious, \\ witch, demons, evil\end{tabular} \\ \hline
\textbf{\begin{tabular}[c]{@{}l@{}}Computationally from \\ the novella\end{tabular}} & \begin{tabular}[c]{@{}l@{}}artist, unconventional, \\ imperturbable, ejaculation, \\ omnibus, inexhaustible, unaffected,\\ incurable, examination, unusually, \\ illness, insane\end{tabular} & \begin{tabular}[c]{@{}l@{}}acceptance, indication, expectation, \\ echo, coincidence, exaggeration, \\ strangeness, excess, renewal, \\ extension, evil, ghost\end{tabular} \\ \hline
\end{tabular}

\caption{Lists of words for the two topics according to each strategy for the construction of the list of words.}
\label{table:topic:word:list}
\end{table*}

\subsection{Approach}

We performed a series of computational experiments to show that the a consistent ambiguity is woven into a novella. In our experiments we used several different word embedding spaces, each of which was used to reflect a different interpretation offered by the text. We also considered temporality, and differentiated between a contemporary reader and a reader at the time of the novella’s publication, by using two embedding spaces (James’ oeuvre and Wikipedia) which broadly represent changes in vocabulary and syntax. We represented each possible interpretation as a list of words reflecting its topic. We estimated the presence of each topic in the text using two different metrics:

\begin{enumerate}
    \item The average cosine similarity between all of the tokens in text and the words in the list representing each topic.
    \item The word mover's distance (WMD) \cite{Kusner:2015:WED:3045118.3045221} between the topic's list of words and the text.
\end{enumerate}

\subsubsection{Topic representation as a list of words}
We employed two strategies to construct the list of words for each topic representation.
\begin{itemize}
    \item \textbf{Strategy 1: Manual extraction}. First, we manually extracted the indicative words from the Wikipedia page that corresponds to the topic. For the insanity topic the words were manually collected from the \textit{insanity} Wikipedia page\footnote{\url{https://en.wikipedia.org/wiki/Insanity}}. For the ghost topic the words were manually collrected from the \textit{ghost} Wikipedia page\footnote{\url{https://en.wikipedia.org/wiki/Ghost}}.
Second, We manually extracted the indicative words from the novella itself. The words were selected by a professional literary researcher.

    \item \textbf{Strategy 2: Computational extraction}. From each embedding space, we extracted the top $k$ cosine-similar words for the topic's $seed$. We defined a topic's seed as a very short list of up to three words corresponding to the very core of the topic and used in classic manual literary analysis. For the insanity topic, we used the words: \textit{insane} and \textit{illness} as the seed words. For the ghost topic we used the words: \textit{ghost} and \textit{evil} as the seed words. We add the words from the topic's seed to the word list as well. The resulting word list is of $k+size(seed)$ length.   
\end{itemize}

The four word lists for each topic are presented in Table \ref{table:topic:word:list}.

\begin{table*}[]
\centering
\begin{tabular}{lllll}
\hline
\textbf{Topic's Source \textbackslash Metric and Embedding Space} & \textbf{cosine avg W} & \textbf{cosine avg J} & \textbf{WMD W} & \textbf{WMD J} \\ \hline
Insanity - manually from Wikipedia & 0.081 & 0.187 & 7.501 & 7.447 \\
Ghost - manually from Wikipedia & 0.045 & 0.145 & 7.680 & 7.564 \\ \hline
Insanity - manually from ToS & 0.099 & \textbf{0.205} & 7.258 & 7.277 \\
Ghost - manually from ToS & 0.067 & \textbf{0.198} & 7.622 & 7.539 \\ \hline
Insanity - computationally from Wikipedia & 0.103 & 0.125 & 7.896 & \textbf{7.538} \\
Ghost - computationally from Wikipedia & 0.088 & 0.170 & 7.789 & \textbf{7.517} \\ \hline
Insanity - computationally from James' oeuvre & 0.039 & 0.203 & 7.896 & 7.613 \\
Ghost - computationally from James' oeuvre & 0.097 & 0.236 & 7.080 & 7.580 \\ \hline

\end{tabular}
\caption{Ambiguity quantification of the insanity and ghost topics and various embedding spaces using two metrics: the average (avg) cosine similarity and WMD between the topic's word list and the text of "The Turn of the Screw" (TOS). Wikipedia embedding space is denoted as W and the complete James' bibliography embedding space is denoted as J; the most obvious ambiguity presence appears in \textbf{bold}.}
\label{table:duality:results}
\end{table*}
\subsubsection{Embedding Spaces}

We used the following corpora based embedding spaces to reflect the perspective of the author,the author's contemporaries, and today's reader:

\begin{enumerate}
    \item GloVe \cite{Pennington14glove:global} pre-trained modern English embeddings that were trained on English Wikipedia and Gigaword. We used this embedding space to reflect today's readers' perception of the words in text.
    \item word2vec \cite{mikolov2013efficient} embeddings trained on the complete bibliography of Henry James. This corpus, which consists of 90 texts, is available from project Gutenberg\footnote{\url{http://www.gutenberg.org/ebooks/author/113}}. We used this embedding space in an attempt to capture James' linguistic regularities of semantic relatedness as they appear in his works. The same embedding space is used to approximate the perception of a reader from the end of the 19th century.
\end{enumerate}

\subsection{Experimental Settings}

We lowercased and tokenized the text using the NLTK \cite{nltk} toolkit. For the complete James' bibliography we used the Gensim \cite{gensim} toolkit to compute the 300-dimensional word2vec \cite{mikolov2013efficient} embeddings. For the Wikipedia embedding we used off-the-shelf pre-trained GloVe embeddings\footnote{\url{https://nlp.stanford.edu/projects/glove/}}. When computing the average cosine similarity and WMD we removed the stop words from the computation; we used NLTK's stop words list. We set k=10 and computed top-10 most similar words to the topic's seed. The complete code is available\footnote{\url{https://github.com/vicmak/TurnOfTheScrew}}.

\subsection{Evaluation}
\label{section:eval}
We computed the average cosine similarity and WMD for each topic's word list representation in the two embedding spaces. The results are presented in Table \ref{table:duality:results}. 

\noindent \textbf{Cosine similarity metric.} When we computed the topic's presence with the average cosine similarity metric the most obvious ambiguity is present when the topic's word list was obtained by an expert's \textit{close reading} of "The Turn of the Screw", measured in the embeddings calculated from the complete James' bibliography. We make two observations: 1) The ratio between the topics presence $\frac{Insanity}{Ghost} = 1.035$ and 2) the degree of cosine similarity which is 0.205 for the insanity topic and 0.198 for the ghost topic, is relatively high. We thus conclude that for this metric, the presence for ambiguity is the most obvious in the context of 19th century vocabulary and syntax. 

\begin{figure*}
  \begin{subfigure}{0.50\textwidth}
    \includegraphics[width=\linewidth]{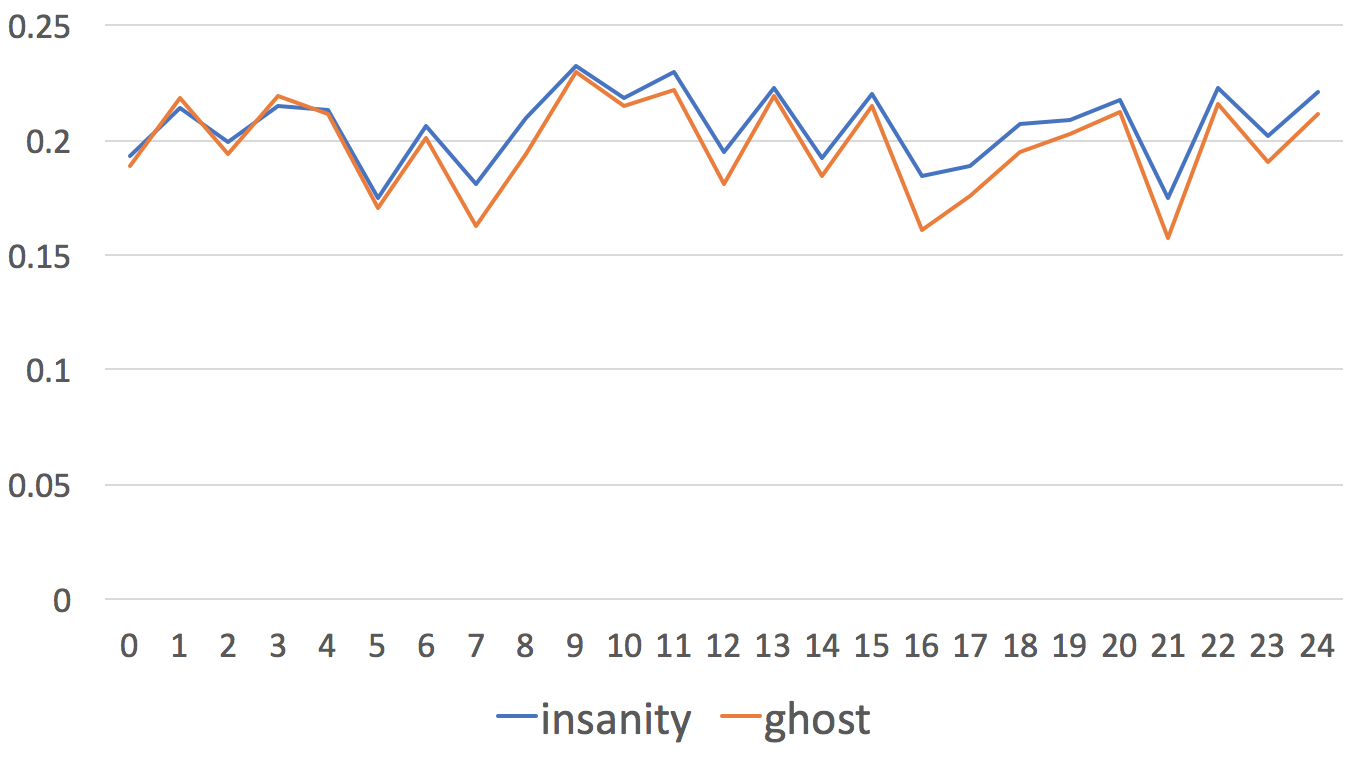}
    \caption{Average cosine similarity of the two topics' word lists for each of the chapters (topic word list selection with strategy-1 by literary expert).} \label{fig:1a}
  \end{subfigure}%
  \hspace*{\fill}   
  \begin{subfigure}{0.50\textwidth}
    \includegraphics[width=\linewidth]{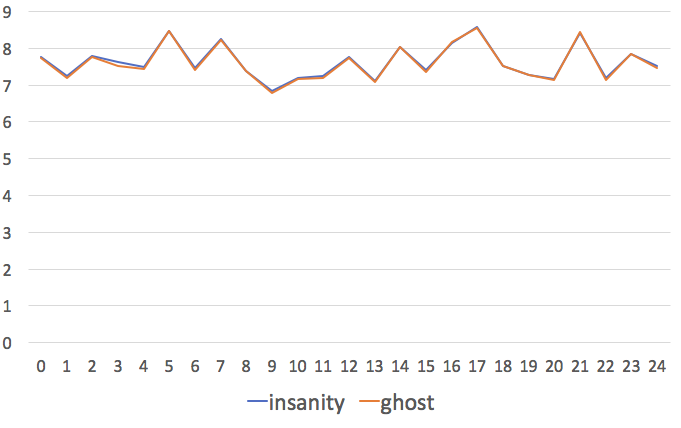}
    \caption{WMD of the two topics' word lists for each of the chapters (topic word list selection with strategy-2 from Wikipedia).} \label{fig:1b}
  \end{subfigure}%
  \hspace*{\fill}   

\caption{Analysis of the novella's ambiguity by chapter (both metrics were calculated in the complete James' bibliography embedding space).} \label{figure:chapters:duality}
\end{figure*}

When choosing the topic's word list with Strategy-2 and computing the average cosine similarity in complete James' bibliography embeddings space the degree of cosine similarity is also high: 0.203 and 0.236 respectively. That is, the two topics are present to a significant degree in the text. However, in this topic representation strategy the ratio between the insanity topic presence ang the ghost topic presence is $\frac{Insanity}{Ghost} = 0.86$, implies a less present ambiguity from the reader's point of view as opposed to the Strategy-1 where topic's word list was composed by an expert. 

In light of these results, we argue that the ability to capture the novella’s implicit ambiguity is best achieved via the combined effort of a human expert and  computational techniques.
 
\noindent \textbf{WMD metric.} When we computed the topic's presence using the WMD metric, the most obvious ambiguity is present when the topic's word list was obtained with Strategy-2, in the complete James' bibiography embedding space. the ratio between the insanity topic presence and ghost topic presence is $\frac{Insanity}{Ghost} = 1.002$. 

Interestingly the lowest presence of ambiguity was observed for both evaluation metrics when the topic's word list was obtained with Strategy-2 in Wikipedia embedding space, and the metrics themselves were computed in this space. We attribute this this result to the fact that \textit{The Turn of the Screw} was written in the late 19th century. Wikipedia's embeddings space's ability to reflect the perception of a 19th century reader, writer or narrator is very limited, as one would expect.

\section{Novella's Content Analysis}
\label{section:content}

\subsection{Ambiguity in the Narration's Progress Analysis}
"The Turn of the Screw" first appeared in 1898 in the New York illustrated magazine, \textit{Collier’s Weekly}, where it was published in 12 installments between January 27 and April 16. Only in October of that year did the story appear in one piece in its entirely in James’ "The Two Magics"; it later appeared in full in volume 12 of his 1908 New York Edition. The current analysis takes into consideration the often-neglected original serialized division of the work into weekly installments in order to examine whether the fluctuations between the story’s two main contested interpretations are punctuated by the original segmentation and chapter order. The analysis importantly demonstrates that the 12 installments and chapters' order set the rhythm of the story’s ambiguity. With few exceptions, each chapter functions as a counter to the previous chapter in terms of its advancement of the \textit{ghost} vs. \textit{insanity} paradigms: when one chapter advances both interpretations intensely, the next chapter decreases that intensity and so forth. 
To perform the computational analysis, we divided the novella in two different ways: 1) according to the 12 originally published installments, and 2) based on the novella's chapters. We computed the metrics based on the progress in the two division options. We used the topics' word lists and embedding space according to the best strategy as described in section \ref{section:eval}.

\begin{figure*}
  \begin{subfigure}{0.50\textwidth}
    \includegraphics[width=\linewidth]{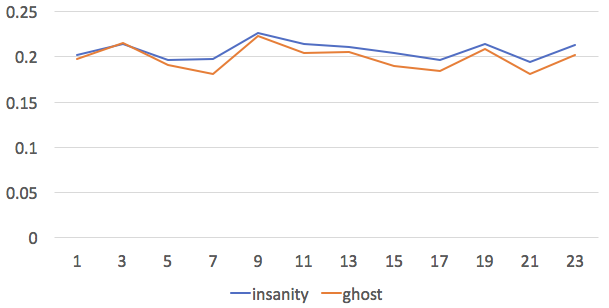}
    \caption{Average cosine similarity of the two topics' word lists to each of the published installments (topic word list selection with strategy-1 by literary expert).} \label{fig:2a}
  \end{subfigure}%
  \hspace*{\fill}   
  \begin{subfigure}{0.50\textwidth}
    \includegraphics[width=\linewidth]{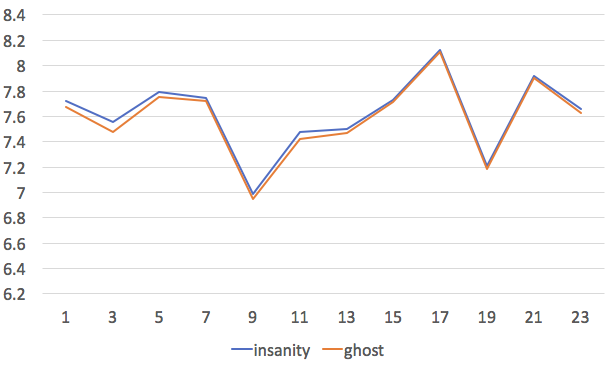}
    \caption{WMD of the two topics' word lists to each of the published installments.  (topic word list selection with strategy-2 from Wikipedia).} \label{fig:2b}
  \end{subfigure}%
  \hspace*{\fill}   
\caption{Analysis of the novella's ambiguity by installment (originally printed by Collier's Weekly in 12 installments), both metrics were calculated in the complete James' bibliography embedding space.} \label{figure:collier:duality}
\end{figure*}

The presence of ambiguity (measured with both metrics) throghout the novella based on its actual division of chapters is presented in Figure \ref{figure:chapters:duality}, while Figure \ref{figure:collier:duality} presents the presence of ambiguity (measured with both metrics) throughout the novella based on the installments of the original serialized version of the novella.
For both metrics (average cosine similarity and WMD), it is clear that the ambiguity is not concentrated in one particular chapter or installment. Not only does the ambiguity appear in all chapters/installments, but the level of each topic's presence changes accordingly. In the case of WMD metric for the Strategy-2 in James' complete bibliography and the same embeddings space, the level of the presence of ambiguity is barely distinguishable in the graphs both in Figure \ref{fig:1b} and in Figure \ref{fig:2b}. 

\subsection{Linguistic Content Analysis}

Literary ambiguity is a complex phenomenon that requires analysis performed by a literary expert to initially spot and then explain its presence. We explored the ability of common NLP techniques to perform a deeper analysis of the novella.

First, to further emphasize the difficulty of initial oberving the ambiguity, we performed a series of LDA \cite{LDA} experiments to gain more insight into the topic analysis of the novella's text. We tried various chapter-topic densities.  The LDA results revealed several topics that are present in the book, most of which are the first names of the novella's characters (e.g., Mrs. Grose) and motifs related to children, none of which was connected to the topics we examined (i.e., insanity and ghost). Furthermore,  none of the topics discovered by LDA explicitly specified the presence of ambiguity. For example consider the following topics we obtained during the experiments:
\begin{itemize}
    \item Topic-1: little, could, never, one, made, might, would, still, well, know
    \item Topic-2: might, would, one, relief, moment, quint, many, ever, enough, strange
\end{itemize}
We omit the complete results of the LDA analysis for brevity.

Second, we tried to learn which words characterize "The Turn Of The Screw" and differentiate it from other novellas written by James. We used the Kullback-Leibrer (KL) divergence \cite{10.1145/312624.312681, kullback1951information} to estimate the relative distinguishable role of the words James used in this novella, as opposed to the words used in his other works by examining those words which contribute the most to the divergence score. After we omit the first names of the novella's characters, we get the following nine words: \textit{pupils, schoolroom, someone, nonetheless, colleague, pool, childish, visitant, naughty}.

Both LDA and KL divergence based analysis support the novella's main narration theme. Words like \textit{strange, visitant, naughty} and \textit{relief} indicate supernatural phenomena, however, do not imply whether this phenomena is real or the fruit of the imagination of a mentally unstable individual.  

\subsection{Punctuation Analysis}

\begin{figure}[]
    \centering
    \includegraphics[width=0.50\textwidth]{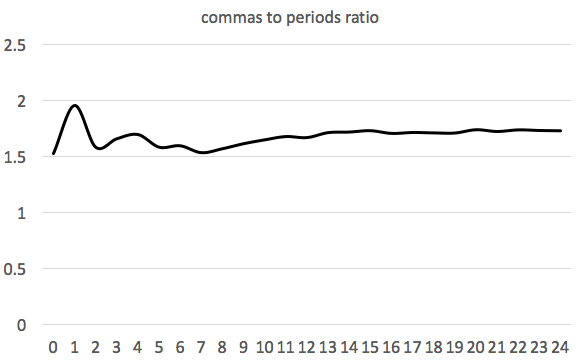}
    \caption{Comma to period ratio by chapter in the novella's text.}
    \label{fig:commas:periods}
\end{figure}

Following the practice of \citet{piper2018enumerations}, we perform punctuation analysis throughout the novella's chapters. We plot the cumulative ratio between commas and periods with the novella's narration by chapter in Figure \ref{fig:commas:periods}. The ratio is the highest (i.e., there is the smallest number of periods compared to the number of commas) in the first chapter where the story line begins and there is a switch of the narrator - Douglas reading the story as it is told by the governess. In the prologue (Chapter 0), the ratio is the lowest, which corresponds to the explanation of \citet{piper2018enumerations} that there are fewer periods at the beginning of the story and more periods towards the story’s end where the narrative begins to converge. Analyzing the plot, we conclude that there is no significant change in the comma to period ratio. This observation supports the presence of a ambiguity that is likely very difficult to detect by a non-expert reader who enjoys the novella, with a clear end.

\section{Discussion}
\label{section:discusion}
The current article is the first of its kind to provide a calculated, visual demonstration of the extent to which "The Turn of the Screw" is indeed meticulously structured around systematic ambiguity. The graphs in Figure \ref{figure:chapters:duality}, a product of the novella's computational analysis, exhibit an exceptional, almost uncanny shared textual fluctuation between the two possible interpretations. The patterns of these visual models, where the two lines beautifully rise and fall in tandem, give a mathematic and visual account of the author's meticulous craftsmanship and the literary sensibility provided by computational analysis when it is thoughtfully used by careful readers. As \citet{piper2018enumerations} recently wrote, “visuality does not simply sit alongside quantity as two equally forgotten dimensions of reading (iconoclasm as arithmophobia’s twin)... the diagram [is] a necessary vehicle that can be used to envision quantity, to grasp, in however mediated a fashion, the quantitative dimension of texts”. 

In the case of James’ novella, the graph gives visual form not only to the internal mechanism of a remarkable literary work, but also gives form to a longstanding critical history, which itself continually shifted from one line of the graph to the other. In addition, this graph, like the work it visualizes, speaks to the process at the heart of literary interpretation. As various critics have demonstrated, \textit{The Turn of the Screw} brings to the fore the readerly effort at the core of the hermeneutic act. In David Bromwich’s \cite{james2011turn} astute words, “The Turn of the Screw has become one of the central modern texts for understanding what interpretation is in literature—the grammar and limits of the perceptual process by which we sort materials for interpretation into evidence on one side and surmise on the other”.

Indeed, that ambiguity is one of the basic attributes of literature has been a premise underlying the discipline of literature from its very foundation by such pioneer schools as the Russian Formalism, Czech Structuralism, New Criticism, and the Tel Aviv School of Poetics and Semiotics \cite{empson2004seven, brooks1947well, perry1986king, jakobson2011sound}, an assumption that, as \citet{ossa2019history} shows, runs as far back as antiquity. The graphs in Figure \ref{figure:chapters:duality}, following the internal movements of the unambiguous text, chart the flexibility continuously demanded from the reader as he/she shifts between the possible interpretations. In other words, these graphs afford us with a visual demonstration of what ambiguity looks like.

\section{Conclusions and Future Work}
\label{section:conclusions}
While most NLP  research that examines ambiguity attempts to resolve it, in this paper we showed how computational methods can \textit{sense}, explain, characterize, and demonstrate subtle ambiguous text. The use of a subject that has been researched and debated for a century allowed us to perform this work and computationally sense a very complex ambiguity along the novella’s narration progress. The novella’s analysis with LDA and KL divergence techniques supports the difficulty of directly observing the dual interpretation. The use of cosine similarity and WMD based on word embeddings from various embedding spaces showed a consistent and rhythmic level of ambiguity across the chapters. The differences in the results obtained from a modern Wikipedia embedding space vs James’ hundred-year-old repertoire emphasize the effect of time on vocabulary and the importance of considering the historical context of a text’s publication in NLP research, as is the norm in the humanities.

The most intriguing direction for future research is examining ambiguous text generation \cite{GANGeneration} using advanced methods such as GAN \cite{GAN} and encoder-decoder \cite{cho2014learning} architectures.

\bibliography{tacl2018}
\bibliographystyle{acl_natbib}

\end{document}

%% file: macros.tex
\newcommand{\vm}[1]{\textcolor{red}{$\ll$\textsf{#1 --VM}$\gg$}}